\pdfoutput=1

\documentclass[11pt]{article}

\usepackage{EACL2023}

\usepackage{times}
\usepackage{latexsym}

\usepackage[T1]{fontenc}

\usepackage[utf8]{inputenc}

\usepackage{microtype}

\usepackage{inconsolata}

\usepackage{graphicx}
\usepackage{amsmath}
\usepackage{algorithm}
\usepackage{algorithmic}
\usepackage{multirow}

\title{Exploiting Summarization Data to Help Text Simplification}

\author{Renliang Sun, Zhixian Yang, Xiaojun Wan \\
  Wangxuan Institute of Computer Technology, Peking University \\
  Center for Data Science, Peking University \\
  The MOE Key Laboratory of Computational Linguistics, Peking University \\
  \texttt{\{sunrenliang, yangzhixian\}@stu.pku.edu.cn} \\
  \texttt{wanxiaojun@pku.edu.cn} \\}

\begin{document}
\maketitle
\begin{abstract}
One of the major problems with text simplification is the lack of high-quality data. The sources of simplification datasets are limited to Wikipedia and Newsela, restricting further development of this field. In this paper, we analyzed the similarity between text summarization and text simplification and exploited summarization data to help simplify. First, we proposed an alignment algorithm to extract sentence pairs from summarization datasets. Then, we designed four attributes to characterize the degree of simplification and proposed a method to filter suitable pairs. We named these pairs Sum4Simp (S4S). Next, we conducted human evaluations to show that S4S is high-quality and compared it with a real simplification dataset. Finally, we conducted experiments to illustrate that the S4S can improve the performance of several mainstream simplification models, especially in low-resource scenarios.
\end{abstract}

\section{Introduction}

Text simplification and text summarization are two major techniques aiming at improving text readability \citep{margarido2008automatic}. 
The main objective of text simplification is to reduce the complexity of the text while keeping its meaning unchanged \citep{alva2020data,al2021automated}. Text summarization is to summarize the main idea of the document in less space \citep{el2021automatic}.

One of the major problems of text simplification is the lack of high-quality aligned data, which is essential for training most simplification models. Existing text simplification datasets are derived from Wikipedia \cite{zhang2017sentence} and Newsela \cite{xu2015problems}. Researchers have proposed various alignment algorithms to extract complex-simple sentence pairs from articles \cite{jiang2020neural}. However, aligning sentences from only two corpora hinders the acquisition of more simplification data, which motivates us to explore new ways to address this problem.

Text simplification usually involves the operations of keeping, deleting, reordering, etc.\cite{xu2016optimizing} Text summarization does not require a summary to be a simple text. 
Nevertheless, when we analyzed the datasets of text summarization meticulously, we noticed that there are many instances where several sentences in the original document are merged into one sentence, and complex parts are rewritten, as shown in Table \ref{tab:example}. Then, a question arises naturally: to what extent is text summarization correlated with text simplification? Furthermore, is it feasible to extract data from text summarization to help low-resource text simplification?

\begin{table}[h]
\centering
\small
\renewcommand\arraystretch{1.12}{
\begin{tabular}{ll}

\hline 
\multicolumn{2}{l}{Example}                                                                                                                                                                                                                                                                                                                        \\  \hline
\multicolumn{1}{c|}{document} & \begin{tabular}[c]{@{}l@{}}What's Hollywood's \textbf{role in all of this?} \textbf{The}\\ \textbf{same as it has always been} -- to make \\ money.\end{tabular}                                                                                                                              \\ \hline
\multicolumn{1}{c|}{summary}  & \begin{tabular}[c]{@{}l@{}}What does Hollywood \textbf{want?} To make \\ money, \textbf{of course.}\end{tabular}                                                                                                                                                                                                             \\ \hline
\end{tabular}}
\caption{The bolded parts indicate that the complex sentence in the document has been rewritten.}
\label{tab:example}
\end{table}

In this study, we investigated the above problems with a three-step procedure: (1) Extract aligned sentence pairs from summarization datasets. (2) Select sentence pairs in which the source sentences have been simplified. (3) Evaluate the quality of these sentence pairs for text simplification.

To extract aligned sentence pairs from the summarization datasets, we proposed an alignment algorithm based on the similarity between sentences. Then, we designed four attributes and a method to filter sentence pairs suitable for text simplification. We performed human evaluations and conducted experiments using mainstream simplification models on these pairs to show that they are of high quality and can help simplification. 

To summarize, our contributions include: (1) We are the first to exploit summarization data to help text simplification, verifying a new source of simplification data. (2) We proposed an alignment algorithm and a method for filtering complex-simple sentence pairs. We named them Sum4Simp (S4S). (3) We performed both empirical analysis and human evaluations on S4S to verify its quality, and the experimental results with several simplification models show the benefits of S4S for text simplification. The S4S dataset and codes are released at \url{https://github.com/RLSNLP/Sum4Simp}.

\section{Related Work}

\subsection{Simplification Models}

Early text simplification models are mainly based on statistic machine learning \cite{wubben2012sentence, kauchak2013improving, narayan2014hybrid}. In recent years, many scholars have proposed models based on deep learning technology, such as NTS\cite{nisioi2017exploring}, DRESS-LS\cite{zhang2017sentence}, EditNTS\cite{dong2019editnts}, ACCESS\cite{martin2020controllable}, which advance the development of text simplification.

\subsection{Mine Data for Simplification}

The above models require a large number of aligned texts for training. Nevertheless, text simplification is a low-resource problem. Some works aim at designing unsupervised models \cite{qiang2019unsupervised,surya2019unsupervised,kumar2020iterative,laban2021keep}. While other works try to mine aligned sentence pairs from more data to help train the models. \citet{martin2020multilingual} proposed unsupervised mining technology to create multi-language simplification corpora automatically. \citet{lu2021unsupervised} used the back-translation approach to construct a large-scale pseudo sentence simplification corpus.

\subsection{Relationship with Text Summarization}

For a long time, studies on text simplification and text summarization have been conducted separately.
Nevertheless, there exist circumstances where complex texts not related to the main idea are removed when summarizing a document, and multiple sentences can be compressed and rewritten into a single sentence. Such a summarization can also be regarded as a simplification. \citet{ma2017semantic} proposed a semantic relevance-based model to improve the results of simplification and summarization. \citet{zaman2020htss} pointed out some similarities between the two tasks and defined the new task of generating simplified summaries. 
Up to now, none of the work has specifically analyzed the relationship between summarization and simplification. It is still worth investigating whether the data from summarization can help simplification.

\section{Mine Sentence Pairs for Simplification from Summarization Datasets}

\begin{figure*}[h]
\centering
\includegraphics[width=14.5cm]{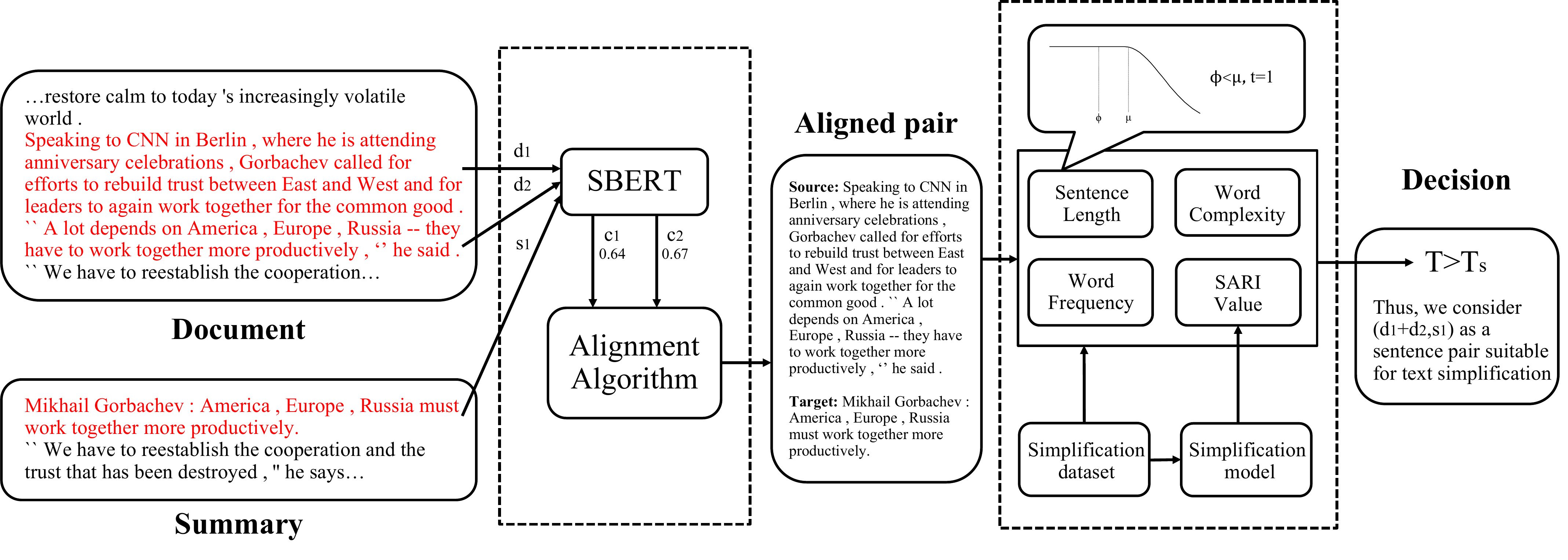} 
\caption{The process of mining suitable sentence pairs from summarization datasets.}
\label{fig:flowchart}
\end{figure*}

In this section, we will elaborate on how to extract sentence pairs that are suitable for text simplification from text summarization datasets. Text summarization is a document-level task while text simplification refers to a sentence-level task. Thus, we proposed an algorithm to extract aligned sentence pairs at first. 
Then, since not all aligned sentence pairs are suitable for text simplification, we chose four attributes and defined a set of rules to filter the appropriate sentence pairs. The whole process is shown in Figure \ref{fig:flowchart}.

\subsection{Sentence Alignment Algorithm}
\label{sec:alignment_algorithm}

Previous sentence alignment algorithms such as CATS \cite{vstajner2018cats} aim at sentence compression (one complex sentence corresponds to one simple sentence) or sentence splitting (a complex sentence is split into several simple sentences). They do not satisfy the requirement to align sentence pairs from summarization datasets, where one sentence in the summary corresponds to multiple sentences in the document. 
Thus, we proposed an alignment algorithm to address this problem.

Assume that there are $m$ sentences in the document and $n$ sentences in the summary. For each sentence $d_i$ in the document and each sentence $s_j$ in the summary, we first compute the similarity between the two sentences. We use SBERT \cite{reimers2019sentence} to achieve this. SBERT is a pre-trained model based on BERT \cite{devlin2019bert}, in which the similarity of two input sentences will be calculated rapidly. Then, we define the upper threshold of similarity $S_{max}$ and the lower threshold of similarity $S_{min}$. $S_{max}$ is greater than $S_{min}$ and they are in the range [0,1]. Assume that the maximum value of similarity between any sentence in the document and $s_j$ is $D_{max}$. If $D_{max}$ is greater than $S_{max}$, we consider that the sentence corresponding to $D_{max}$ is very similar to $s_j$. Therefore, we keep $s_j$ as the target sentence and the sentence corresponding to $D_{max}$ as the source sentence, and they form an aligned sentence pair. If $D_{max}$ is smaller than $S_{min}$, we consider that there is no sentence in the document that is similar to $s_j$. Thus, we do not keep sentence pairs related to $s_j$.

\begin{algorithm}
\renewcommand{\algorithmicensure}{\textbf{Output:}}
\small
\caption{Sentence alignment algorithm}
\label{alg:alignment}
\begin{algorithmic}[1]
\STATE Initialization: F and C are empty sets
\FOR{$d_i$ \textbf{in} $d_1$,$d_2$,...,$d_n$}
\STATE $c_i$ = SBERT($d_i$,$s_j$)
\STATE C.append($c_i$)
\ENDFOR
\IF{max(C)>$S_{max}$}
\STATE F.append(corresponding $d_i$ of max(C))
\ELSIF{$S_{max}$>max(C)>$S_{min}$}
\STATE F.append(corresponding $d_i$ of max(C))
\STATE C.remove(max(C))
\REPEAT
\STATE $c_i$ = SBERT(stitch(F,corresponding $d_i$ of max(C)),$s_j$)
\IF{$c_i$>$S_{add}$}
\STATE F.append(corresponding $d_i$ of max(C))
\STATE C.remove(max(C))
\ENDIF
\UNTIL{$c_i \leq S_{add}$ \textbf{or} len(C)$\geq L_{max}$}
\ENDIF
\ENSURE (F,$s_j$) as an aligned sentence pair
\end{algorithmic}
\end{algorithm}

If $D_{max}$ is greater than $S_{min}$ and smaller than $S_{max}$, we consider this to be the case where multiple sentences in the document correspond to $s_j$. We temporarily save the sentences corresponding to $D_{max}$, and then find the sentence with the largest similarity among the remaining sentences of the document. We stitch this sentence with the sentence we just saved according to the order of the sentences in the document. We repeat this operation until the similarity between the stitched sentences and $s_j$ is less than a threshold. We define this threshold as $S_{add}$, which takes values in the range [$S_{min}$,$S_{max}$]. To prevent the problem of imbalance where the length of the source sentence far exceeds the length of the target sentence caused by extracting too many sentences from the document, we set $L_{max}$. When the number of stitched sentences reaches $L_{max}$, we save these stitched sentences as source sentences and $s_j$ as the target sentence.

\subsection{Four Attributes to Characterize Simplification}
\label{sec:four_attributes}

Aligned sentence pairs obtained from Algorithm \ref{alg:alignment} are not always complex-simple ones, and an example is given below:

\noindent \textbf{Source sentence}: Analysts say the Arab Spring has made Dubai a safe haven for people in the Middle East who worry about the turmoil elsewhere.

\noindent \textbf{Target sentence}: Analysts say the Arab Spring has made Dubai a safe haven for those who worry about the turmoil elsewhere.

This example is a real sentence pair mined from the summarization data. It is an aligned sentence pair but neither the attributive clause nor the complex words such as ``turmoil'' are simplified. Thus, it is not a good instance for text simplification. We design four attributes to characterize whether the source sentence is simplified or not, which are:

\noindent \textbf{Sentence Length} Intuitively, the longer the sentence, the more complex the sentence is likely to be.
We calculate the length of the target sentence minus the average length of the source sentences.

\noindent \textbf{Word Complexity} 
We believe that the lower the average complexity of words, the simpler the sentence. We use a lexicon of word complexity created by \citet{maddela2018word}. Each word is scored by humans. The higher the score, the more complex the word. We calculate the value of the average word complexity of the target sentence minus the average word complexity of the source sentences.

\noindent \textbf{Word Frequency} Some words appear more frequently in complex sentences, while some words appear more frequently in simple sentences. The more frequently a word appears in a simple sentence, the more likely it is to be a simple one. We calculate the odds ratio \cite{monroe2008fightin} to represent the frequency of word occurrence. For two corpus, namely $i$ and $j$, their sizes are $n_i$ and $n_j$, respectively. For a word $w$, the occurrences in corpus $i$ and corpus $j$ are $w_i$ and $w_j$, respectively. Then, the odds ratio $r$ of word $w$ between corpus $i$ and corpus $j$ can be defined as:

\begin{equation}
    r = \frac{w_i/w_j}{n_i/n_j}
\label{odds_ratio}
\end{equation}

We use the simplification dataset to construct a dictionary containing the odds ratios of the words. 
For example, if we want to conduct experiments on WikiLarge \cite{zhang2017sentence}, we calculate the odds ratio of the words occurring in the WikiLarge training set. We calculate the value of the average odds ratio of the target sentence minus the average odds ratio of the source sentence.

\noindent \textbf{SARI Value} SARI \cite{xu2016optimizing} is an essential evaluation method for text simplification. It takes the original sentence, the simplified sentence, and reference sentences into consideration. The SARI value is an average of F1 scores of add and keep operation and precision of delete operation. The score for each operation is obtained by averaging $n$-gram scores.

\begin{equation}
\begin{aligned}
    SARI & = \frac{1}{3}F_{add} + \frac{1}{3}F_{keep} + \frac{1}{3}P_{del} \\
    P_{operation} & = \frac{1}{4}\sum_{n=1,2,3,4}p_{operation}(n) \\
    R_{operation} & = \frac{1}{4}\sum_{n=1,2,3,4}r_{operation}(n) \\
    F_{operation} & = \frac{2\times P_{operation}\times R_{operation}}{P_{operation}+R_{operation}} \\
    ope&ration \in [add,keep,del]
\end{aligned}
\end{equation}

\begin{figure*}[h]
\centering
\includegraphics[width=12cm]{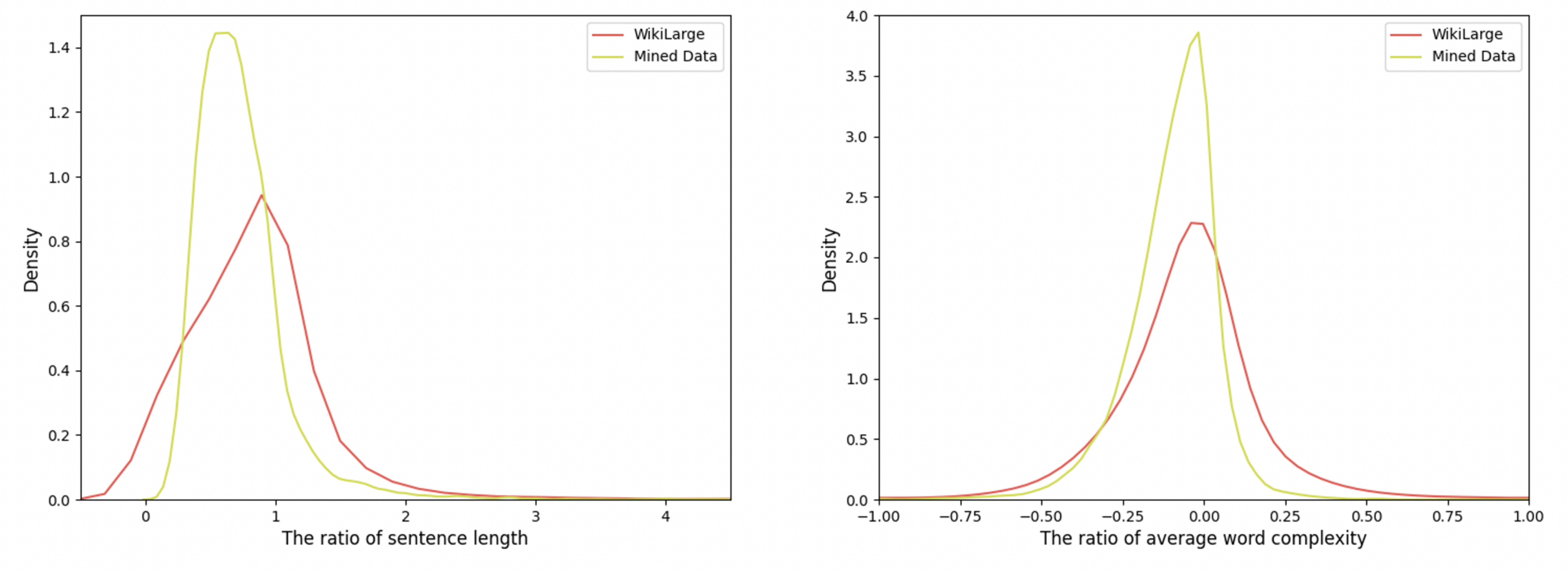} 
\caption{Distributions of the ratio of sentence length and average word complexity. We smoothed the results by using a Gaussian kernel. Sentences from S4S are more compressed than in WikiLarge. Sentences where the words become more complex are also less than in WikiLarge.}
\label{comparison}
\end{figure*}

We consider the source sentence of the aligned sentence pairs as the original sentence and the target sentence as the simplified sentence. We need to train a simplification model at first. 
For example, we trained a model like ACCESS \cite{martin2020controllable} on the WikiLarge training set. 
Then, we input the source sentences into the simplification model and generate simplified sentences. These simplified sentences are used as reference sentences. Finally, the SARI values are calculated.


\subsection{Quantify Simplicity and Filter Suitable Sentence Pairs}
\label{sec:filtering_method}

For each attribute, we propose a method to quantify the simplicity of a sentence. Our method is based on a hypothesis: a reference simplification dataset performs approximately normally distributed on each attribute. Simplification datasets can contain hundreds of thousands of instances, in line with the concept of large samples in statistics. Therefore, we believe this hypothesis is reasonable.

Take the sentence length attribute as an example. We first calculate the mean $\mu$ and standard deviation $\sigma$ of the sentence length of the training set of a reference dataset (e.g. WikiLarge). For a random variable X, the probability density function $f(x)$ can be obtained.
If the ratio of sentence length for a sentence pair is $\phi$, its score $t$ on this attribute is:

\begin{equation}t=\left\{
    \begin{array}{lr}
        1, & \phi<=\mu \\
        2\times(0.5-\int_{\mu}^{\phi}f(x)dx), & \phi>\mu
    \end{array}\right.
\label{eq:three_attributes}
\end{equation}

\begin{equation}
    t=\left\{
    \begin{array}{lr}
    2\times(0.5-\int_{\phi}^{\mu}f(x)dx), & \phi<\mu\\
    1, & \phi>=\mu
    \end{array}\right.
\label{eq:sari_attribute}
\end{equation}

\begin{equation}
    f(x) = \frac{1}{\sqrt{2\pi}\sigma}exp\left(-\frac{(x-\mu)^2}{2\sigma^2} \right)
\end{equation}

The mathematical significance is that if $\phi<=\mu$, the simplification degree of the sentence pair is greater than the average simplification degree of the simplification dataset on this attribute. Thus, we give a score of 1 to $t$. If $\phi>\mu$, we subtract the proportion of sentence pairs with a ratio greater than $\mu$ and lower than $\phi$ that is in the simplification dataset. Then, we perform a normalization operation to obtain $t$.
For attributes sentence length (len), word complexity (comp), and word frequency (freq), a lower $\phi$ indicates a greater degree of simplification. We use Equation (\ref{eq:three_attributes}) to calculate $t$. For attribute SARI value (sari), a higher $\phi$ indicates a greater degree of simplification. We use Equation (\ref{eq:sari_attribute}) to calculate $t$.

To make a final decision, the scores on each attribute are weighted with $\alpha$ and summed to obtain T for a sentence pair, indicating the extent of simplification of the source sentence. We set a threshold value ${\rm T_{s}}$ to control the extent of simplification. When T>${\rm T_{s}}$, we consider the sentence pair suitable for the task of text simplification.

\begin{equation}
\begin{aligned}
    {\rm T} = &\sum_{i\in Attr}\alpha_{i}t_{i} \\
    Attr = [len, &comp, freq, sari]
\end{aligned}
\end{equation}

We exploit and filter sentence pairs from the CNN/Daily Mail summarization dataset \cite{nallapati2016abstractive}, which contains more than 300,000 documents and corresponding summaries from news stories in CNN and Daily Mail. We name these mined sentence pairs Sum4Simp (S4S).

\section{Quantitative Analysis}

In this section, we want to show that Sum4Simp (S4S) is high-quality. We conducted two human evaluations and performed statistics on S4S, comparing it with real simplification datasets.

\subsection{Human Evaluations}
\label{sec:alignment}

First, we want to evaluate the alignment quality of the sentence pairs obtained in Section 3.1. Following \citet{hwang2015aligning}, we defined the quality of alignment into four classes: Good, Good partial, Partial, and Bad. Due to the space limit, details and examples are demonstrated in Table \ref{tab:defination_of_alignment_quality}.

We randomly selected sentence pairs from the aligned pairs obtained by our proposed alignment algorithm. Then, we designed a baseline that does not use our proposed alignment algorithm. When the similarity calculated by SBERT between a sentence in the document and a sentence in the summary is greater than 0.6, we kept this sentence in the document.
As we introduced in Section \ref{sec:alignment_algorithm}, the CATS method \cite{vstajner2018cats} may not be suitable for aligning sentence pairs from summarization datasets. However, we used it as a baseline. 

\begin{figure}[h]
\centering
\includegraphics[width=6.5cm]{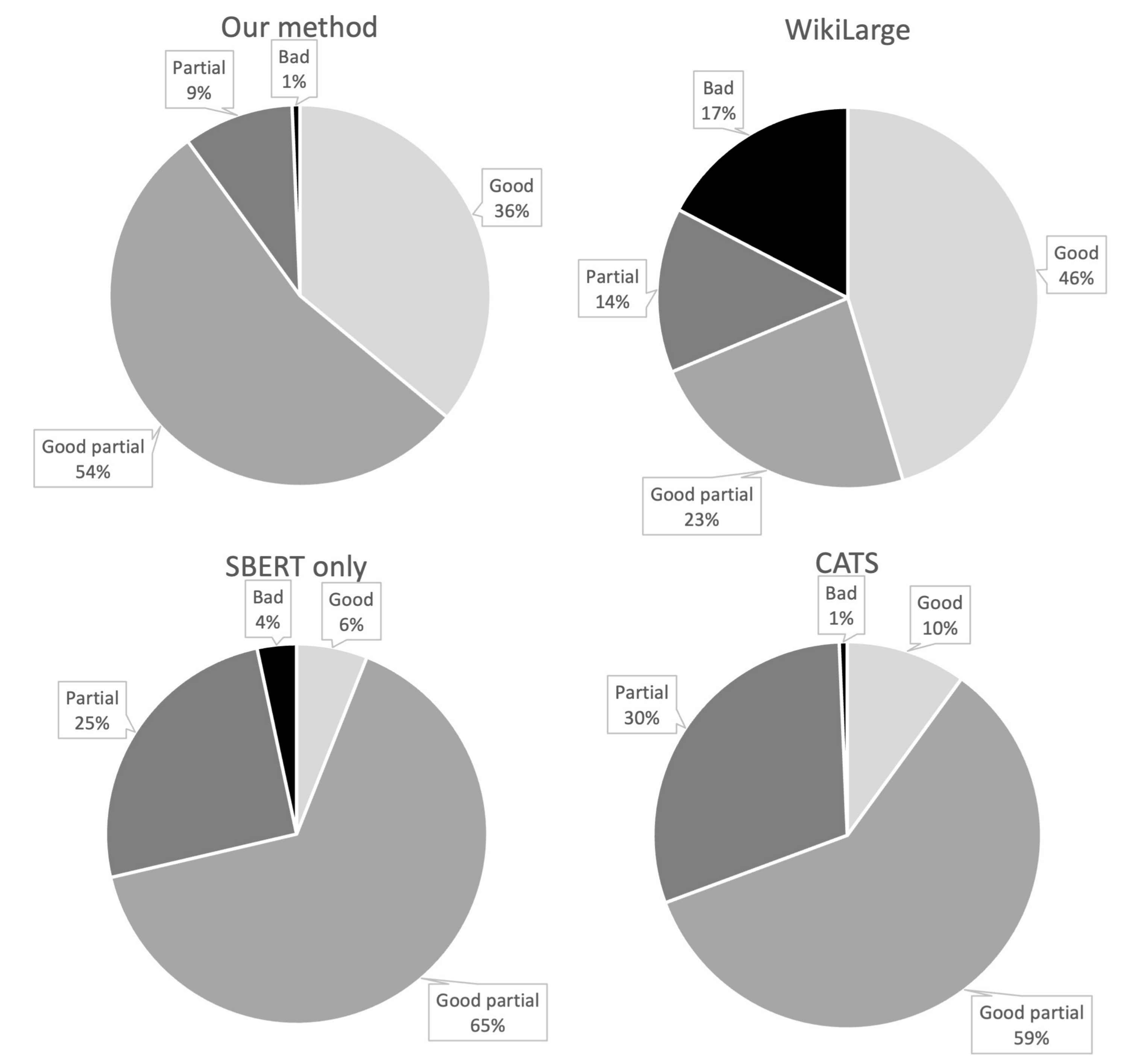} 
\caption{Human evaluation results of data obtained by three alignment methods and WikiLarge. We randomly selected 50 sentence pairs from each source of data. Then, we hired three workers to evaluate the 200 sentence pairs individually.}
\label{pic:alignment}
\end{figure}

We used the two baseline methods described above to obtain aligned sentence pairs from summarization datasets.
What's more, we randomly selected sentence pairs from a simplification dataset named WikiLarge \cite{zhang2017sentence} for comparison. The results are shown in Figure \ref{pic:alignment}.

We considered \textbf{Good} and \textbf{Good partial} to be acceptable quality. The sentence pairs obtained by our proposed alignment algorithm have the highest percentage in these two levels. While WikiLarge has the most sentence pairs with a Good level, it also has the most sentence pairs with a Bad level. \citet{xu2015problems} pointed out that data mined from Wikipedia is not always of high quality.


Then, we want to show that the final sentence pairs obtained in Section \ref{sec:filtering_method} are more suitable for simplification. We randomly selected 50 sentence pairs that are only aligned and 50 sentence pairs from S4S. We also randomly selected 50 sentence pairs from WikiLarge for comparison.

Following \citet{dong2019editnts}, we used two indicators as the criteria: (1) \textbf{Simplicity}: Is the target sentence simpler than the source sentence? (2) \textbf{Adequacy}: Are the source sentence and target sentence fluent and grammatically correct? Another indicator, Meaning, can be regarded as the evaluation of alignment quality, so we did not repeat it. The results are shown in Table \ref{tab:human_evaluation_simplification}. The sentence pairs from S4S receive the highest Simplicity score, significantly higher than the aligned-only pairs and WikiLarge, indicating the effectiveness of the proposed filtering method. 

\begin{table}[h]
\centering
\small
\renewcommand\arraystretch{1.12}{
\begin{tabular}{l|cc}
\hline
                     & Simplicity$\uparrow$    & Adequacy$\uparrow$       \\ \hline
WikiLarge            & 3.11**          & 4.6**           \\
Aligned only         & 3.2**           & 4.81          \\ \hline
S4S & \textbf{3.49} & \textbf{4.94} \\ \hline
\end{tabular}}
\caption{Human evaluation results of data obtained by two methods and WikiLarge. We hired three workers to evaluate individually. Student t-tests were performed and results significantly different from S4S were marked with **(p<0.01).}
\label{tab:human_evaluation_simplification}
\end{table}


\subsection{Statistics and Comparison}

We used three dimensions, sentence length, average word complexity, and odds ratio of cue words, to compare the sentence pairs from S4S with those from WikiLarge. The ratio of sentence length is calculated by dividing the length of the simplified sentence by the length of the original sentence. The ratio of average word complexity is calculated by subtracting the average word complexity of the original sentence from the average word complexity of the simplified sentence.

We randomly selected 10,000 sentence pairs from WikiLarge and S4S, respectively. From Figure \ref{comparison}, in S4S, the number of sentence pairs with a length ratio greater than one has been significantly decreased compared to WikiLarge, indicating that sentences are more compressed. What's more, the vast majority of the ratios of average word complexity are less than zero, suggesting a general simplification at the word level in S4S.

Sentence splitting, a common operation in text simplification, can be represented by the odds ratio of conjunctions and cue words \cite{siddharthan2003preserving}. The definition of the odds ratio is detailed in Equation (\ref{odds_ratio}). When the odds ratio of conjunctions is much less than 1, and the odds ratio of cue words is much greater than 1, a complete degree of simplification is involved. Following \citet{xu2015problems} and \citet{sun2021document}, we calculated the odds ratio of conjunctions and cue words in WikiLarge and S4S, as shown in Table \ref{tab:cue_words_and_conjunctions}.

\begin{table}[h]
\centering
\small
\begin{tabular}{cccc}
\hline
\multicolumn{2}{|c|}{WikiLarge}                                    & \multicolumn{2}{c|}{S4S}                                    \\
\multicolumn{1}{|c}{cue words}    & \multicolumn{1}{c|}{odds ratio$\uparrow$} & cue words                      & \multicolumn{1}{c|}{odds ratio$\uparrow$}    \\ \hline
\multicolumn{1}{|c|}{also}       & \multicolumn{1}{c|}{1.15}       & \multicolumn{1}{c|}{also}     & \multicolumn{1}{c|}{1.13}          \\
\multicolumn{1}{|c|}{then}       & \multicolumn{1}{c|}{1.16}       & \multicolumn{1}{c|}{then}     & \multicolumn{1}{c|}{\textbf{1.21}} \\
\multicolumn{1}{|c|}{still}      & \multicolumn{1}{c|}{1.01}       & \multicolumn{1}{c|}{still}    & \multicolumn{1}{c|}{\textbf{1.41}} \\ \hline
\multicolumn{1}{l}{}             & \multicolumn{1}{l}{}            & \multicolumn{1}{l}{}          & \multicolumn{1}{l}{}               \\ \hline
\multicolumn{2}{|c|}{Wikilarge}                                    & \multicolumn{2}{c|}{S4S}                                    \\
\multicolumn{1}{|c}{conjunctions} & \multicolumn{1}{c|}{odds ratio$\downarrow$} & conjunctions                   & \multicolumn{1}{c|}{odds ratio$\downarrow$}    \\ \hline
\multicolumn{1}{|c|}{and}        & \multicolumn{1}{c|}{0.87}       & \multicolumn{1}{c|}{and}      & \multicolumn{1}{c|}{0.95}          \\
\multicolumn{1}{|c|}{as}         & \multicolumn{1}{c|}{0.72}       & \multicolumn{1}{c|}{as}       & \multicolumn{1}{c|}{0.80}          \\
\multicolumn{1}{|c|}{since}      & \multicolumn{1}{c|}{1.01}       & \multicolumn{1}{c|}{since}    & \multicolumn{1}{c|}{\textbf{0.96}} \\
\multicolumn{1}{|c|}{because}    & \multicolumn{1}{c|}{2.59}       & \multicolumn{1}{c|}{because}  & \multicolumn{1}{c|}{\textbf{1.05}} \\
\multicolumn{1}{|c|}{when}       & \multicolumn{1}{c|}{1.32}       & \multicolumn{1}{c|}{when}     & \multicolumn{1}{c|}{\textbf{1.09}} \\
\multicolumn{1}{|c|}{if}         & \multicolumn{1}{c|}{1.30}       & \multicolumn{1}{c|}{if}       & \multicolumn{1}{c|}{1.38}          \\
\multicolumn{1}{|c|}{but}        & \multicolumn{1}{c|}{1.18}       & \multicolumn{1}{c|}{but}      & \multicolumn{1}{c|}{\textbf{1.11}} \\
\multicolumn{1}{|c|}{though}     & \multicolumn{1}{c|}{0.71}       & \multicolumn{1}{c|}{though}   & \multicolumn{1}{c|}{\textbf{0.62}} \\
\multicolumn{1}{|c|}{although}   & \multicolumn{1}{c|}{0.46}       & \multicolumn{1}{c|}{although} & \multicolumn{1}{c|}{\textbf{0.40}} \\ \hline
\end{tabular}
\caption{The odds ratio of cue words and conjunctions. The bolded parts indicate that S4S performs better than WikiLarge. Some words, such as ``hence'', occur too infrequently to be statistically meaningful.}
\label{tab:cue_words_and_conjunctions}
\end{table}

\section{Experimental Setup}

\subsection{Datasets}

We used two commonly used simplification datasets, \textbf{WikiLarge} \cite{zhang2017sentence} and \textbf{WikiSmall} \cite{zhu2010monolingual}, to demonstrate the usefulness of the sentence pairs mined from summarization data. The training set of WikiLarge contains more than 296k sentence pairs, which is larger than that of WikiSmall containing 88k sentence pairs. We used \textbf{Turkcorpus} \cite{xu2016optimizing} as the validation and the test set for WikiLarge. Each of the 2000 validation instances and the 359 test instances has 8 reference sentences. We used the original validation set and test set for WikiSmall, with 205 validation instances and 100 test instances.


\begin{table*}[h]
\centering
\resizebox{\textwidth}{!}{
\renewcommand\arraystretch{1.2}{
\begin{tabular}{lcccccccccccccccc}
\hline
\multicolumn{1}{|l|}{\multirow{2}{*}{Models}} & \multicolumn{4}{c|}{WikiLarge}                                                                              & \multicolumn{4}{c|}{S4S}                                                                            & \multicolumn{4}{c|}{WikiLarge+OA}                                                                            & \multicolumn{4}{c|}{WikiLarge+S4S}                                                                          \\ \cline{2-17} 
\multicolumn{1}{|l|}{}                           & SARI$\uparrow$       & $F_{keep}$           & $P_{delete}$         & \multicolumn{1}{c|}{$F_{add}$} & SARI$\uparrow$       & $F_{keep}$           & $P_{delete}$         & \multicolumn{1}{c|}{$F_{add}$} & SARI$\uparrow$       & $F_{keep}$           & $P_{delete}$         & \multicolumn{1}{c|}{$F_{add}$} & SARI$\uparrow$       & $F_{keep}$           & $P_{delete}$         & \multicolumn{1}{c|}{$F_{add}$} \\ \hline
\multicolumn{1}{|l|}{Transformer}                & 36.95*                & 70.80                & 36.91                & \multicolumn{1}{c|}{3.15}      & 34.43**                & 58.54                & 43.68                & \multicolumn{1}{c|}{1.08}      & 36.75*                & 70.79                & 36.38              & \multicolumn{1}{c|}{3.06}      & \textbf{37.85}                & 71.11                & 39.15                & \multicolumn{1}{c|}{3.27}      \\
\multicolumn{1}{|l|}{BART}                       & 37.99**                & 72.53                & 37.85                & \multicolumn{1}{c|}{3.59}      & 36.21**                & 64.70                & 42.60                & \multicolumn{1}{c|}{1.34}      & 37.71**                & 73.02                & 36.81                & \multicolumn{1}{c|}{3.31}      & \textbf{39.20}                & 70.99                & 42.31                & \multicolumn{1}{c|}{4.30}      \\
\multicolumn{1}{|l|}{ACCESS}                     & 39.67*                & 71.20                & 42.69                & \multicolumn{1}{c|}{5.12}      & 36.20**                & 65.62                & 41.53                & \multicolumn{1}{c|}{1.44}      & 39.46*                & 69.39                & 43.96                & \multicolumn{1}{c|}{5.03}      & \textbf{40.71}                & 71.26                & 44.06                & \multicolumn{1}{c|}{6.81}      \\ \hline
                                                 & \multicolumn{1}{l}{} & \multicolumn{1}{l}{} & \multicolumn{1}{l}{} & \multicolumn{1}{l}{}           & \multicolumn{1}{l}{} & \multicolumn{1}{l}{} & \multicolumn{1}{l}{} & \multicolumn{1}{l}{}           & \multicolumn{1}{l}{} & \multicolumn{1}{l}{} & \multicolumn{1}{l}{} & \multicolumn{1}{l}{}           & \multicolumn{1}{l}{} & \multicolumn{1}{l}{} & \multicolumn{1}{l}{} & \multicolumn{1}{l}{}           \\ \hline
\multicolumn{1}{|l|}{\multirow{2}{*}{Models}} & \multicolumn{4}{c|}{WikiSmall}                                                                              & \multicolumn{4}{c|}{S4S}                                                                            & \multicolumn{4}{c|}{WikiSmall+OA}                                                                            & \multicolumn{4}{c|}{WikiSmall+S4S}                                                                          \\ \cline{2-17} 
\multicolumn{1}{|l|}{}                           & SARI$\uparrow$       & $F_{keep}$           & $P_{delete}$         & \multicolumn{1}{c|}{$F_{add}$} & SARI$\uparrow$       & $F_{keep}$           & $P_{delete}$         & \multicolumn{1}{c|}{$F_{add}$} & SARI$\uparrow$       & $F_{keep}$           & $P_{delete}$         & \multicolumn{1}{c|}{$F_{add}$} & SARI$\uparrow$       & $F_{keep}$           & $P_{delete}$         & \multicolumn{1}{c|}{$F_{add}$} \\ \hline
\multicolumn{1}{|l|}{Transformer}                & 36.35*                & 66.69                & 40.53                & \multicolumn{1}{c|}{1.82}      & 36.75                & 60.23                & 49.49                & \multicolumn{1}{c|}{0.53}      & 36.38*                & 64.46                & 40.54                & \multicolumn{1}{c|}{4.15}      & \textbf{38.57}                & 66.56                & 43.69                & \multicolumn{1}{c|}{5.46}      \\
\multicolumn{1}{|l|}{BART}                       & 35.13*                & 64.94                & 35.86                & \multicolumn{1}{c|}{4.59}      & 34.13*                & 61.06                & 39.95                & \multicolumn{1}{c|}{1.39}      & 34.65*                & 67.09                & 31.92                & \multicolumn{1}{c|}{4.93}      & \textbf{36.58}                & 67.39                & 37.14                & \multicolumn{1}{c|}{5.22}      \\
\multicolumn{1}{|l|}{ACCESS}                     & 35.35*                & 65.01                & 38.50                & \multicolumn{1}{c|}{2.53}      & 34.63**                & 51.07                & 51.76                & \multicolumn{1}{c|}{1.05}      & 35.67*                & 60.95                & 44.29                & \multicolumn{1}{c|}{1.77}      & \textbf{38.28}                & 58.45                & 53.64                & \multicolumn{1}{c|}{2.73}      \\ \hline
\end{tabular}}
}
\caption{Results of three simplification models trained on four different training sets. The test sets in the upper and lower tables are Turkcorpus and WikiSmall, respectively. ``+'' represents the operation to mix the two datasets and sort them randomly. OA is a set of sentence pairs with a similar size to S4S drawn from aligned but not filtered sentence pairs. The bolded part indicates the training set that achieves the best result for each model. Student t-tests were performed, and SARI values that were significantly different from WikiLarge+S4S and WikiSmall+S4S were marked with *(p<0.05) or **(p<0.01).}
\label{tab:expand_datasets}
\end{table*}

\subsection{Evaluation Metrics and Models}

We took \textbf{SARI} \cite{xu2016optimizing} and \textbf{BERTScore} \cite{zhang2019bertscore} as the evaluation metric in this paper. SARI is the most popular automatic evaluation metric for text simplification. The SARI value is obtained by averaging the $F_{keep}$, $P_{delete}$, and $F_{add}$ score. We used the \textbf{EASSE} package \cite{alva2019easse} to get SARI values. A recent study recommends using BERTScore$_{precision}$ to evaluate the quality of the system outputs prior to using SARI to measure simplification \cite{alva2021suitability}. FKGL \cite{kincaid1975derivation} was used to measure text readability but was proven to be inappropriate for evaluating text simplification recently \cite{tanprasert2021flesch}. BLEU \cite{papineni2002bleu} has been proven to be unsuitable for evaluating text simplification \cite{sulem2018bleu}. Therefore, we did not report FKGL values and BLEU values.

We selected three representative models - \textbf{Transformer} \cite{vaswani2017attention}, \textbf{BART} \cite{lewis2020bart}, and \textbf{ACCESS} \cite{martin2020controllable} to conduct experiments. Transformer and BART perform strongly for many generation tasks.  ACCESS is a simplification model proposed recently and it uses explicit tokens related to different attributes to control the process of simplification. 

\subsection{Training Details}
\label{detail}



We used the Huggingface Transformers \cite{wolf2020transformers} to implement the Transformer model and the BART model. We used the original code to implement the ACCESS model. We used four Nvidia A40 GPUs for training. We reported the results of the model on the test set which has the best SARI value on the validation set.

More details can be found in Appendix \ref{sec:appendix}.

\section{Experimental Results}

\begin{table*}[h]
\centering
\resizebox{12.5cm}{!}{
\renewcommand\arraystretch{1.2}{
\begin{tabular}{|l|cccc|cccc|cccc|}
\hline
\multirow{2}{*}{Models} & \multicolumn{4}{c|}{S4S}                               & \multicolumn{4}{c|}{WikiLarge}                                & \multicolumn{4}{c|}{S4S+WikiLarge}                            \\ \cline{2-13} 
                  & SARI$\uparrow$ & $F_{keep}$ & $P_{delete}$ & $F_{add}$ & SARI$\uparrow$ & $F_{keep}$ & $P_{delete}$ & $F_{add}$ & SARI$\uparrow$ & $F_{keep}$ & $P_{delete}$ & $F_{add}$ \\ \hline
Transformer       & \textbf{44.75}          & 53.32      & 74.72        & 6.19      & 32.59          & 45.38      & 51.78        & 0.61      & 43.61          & 52.24      & 73.91        & 4.68      \\
BART              & 46.42          & 57.20      & 76.62        & 5.43      & 32.98          & 47.12      & 50.10        & 1.70      & \textbf{46.51}          & 57.24      & 73.91        & 4.68      \\
ACCESS            & \textbf{40.19}          & 45.85      & 72.82        & 1.88      & 30.10          & 44.30      & 43.99        & 2.01      & 38.45          & 43.35      & 70.71        & 1.30      \\ \hline
\end{tabular}}
}
\caption{Results on three simplification models trained on three different training sets. The valid and test sets come from S4S.}
\label{tab:cross}
\end{table*}

\subsection{Results on Existing Test Sets}
\label{sec:result_exist}

We designed four types of training sets and tested the three simplification models on existing test sets. We first measured the outputs of each model using BERTScore$_{precision}$ and found that the values are very close to 1, indicating that the outputs are of high quality. Then, the SARI values are shown in Table \ref{tab:expand_datasets}.

From the upper table, Sum4Simp (S4S) mixed with the WikiLarge training set improves the performance of all three simplification models on Turkcorpus. To be more specific, in terms of the SARI metric, ACCESS is improved by 1.04 points, BART is improved by 1.21 points, and Transformer is improved by 0.90 points. We have used the original codes and followed the original hyper-parameter settings, but the SARI value of the ACCESS model trained on WikiLarge is lower than the results reported by \citet{martin2020controllable}. We think this is because we lowered the training data and used the NLTK package to split the words. Meanwhile, seen from the lower table, S4S mixed with the WikiSmall training set also improves the performance of all three models on the test set of WikiSmall. The improvement on the WikiSmall test set is more significant than that on the Turkcorpus test set. In terms of the SARI metric, ACCESS is improved by 2.93 points, BART is improved by 1.45 points, and Transformer is improved by 2.22 points. Example outputs are given in Table \ref{tab:Example1}. It may seem strange that the SARI value of Transformer is higher than that of BART. However, we noticed that the SARI value of BART is approximately 3 points higher than that of Transformer on the validation set, making the experimental results remain convincing.

The size of the training set of WikiLarge is much larger than that of WikiSmall. Therefore, the models were more fully trained on WikiLarge. While the size of the training set of WikiSmall is comparatively smaller, S4S helps the model learn to simplify sentences better and results in a more significant improvement.

OA was designed to verify that the improvement of the results comes from high-quality mined sentence pairs rather than mere data expansion. Compared with the original training set, the performances on WikiLarge+OA and WikiSmall+OA were not improved and even dropped for the model like BART. The results illustrate that the method for filtering suitable sentence pairs for simplification purposes is essential.

If we only used S4S as the training set, the SARI values obtained are 2.5 points lower than the model trained with WikiLarge and 0.5 points lower than the model trained with WikiSmall on average. We believe the performance gap is due to domain differences: S4S comes from news stories written by professional journalists, while WikiLarge and WikiSmall come from Wikipedia. Overall, though S4S comes from a different domain, it can still be beneficial to the existing simplification datasets. 

\subsection{Results on S4S Test Set}
\label{sec:real_simp_dataset}

In this subsection, we treat S4S as a standard simplification dataset that contains more than 243K sentence pairs. We divided the train/dev/test set as 240k/2k/1k, respectively. We would like to see the performance of simplification models on the S4S dataset and we want to know if the WikiLarge dataset from a different domain can improve the performance. We designed three types of training sets. Then, we conducted experiments with each of them to train the three simplification models.

According to Table \ref{tab:cross}, all three simplification models trained on the S4S dataset have significantly higher SARI values compared to the results in Table \ref{tab:expand_datasets}. When we mixed the training set of S4S and WikiLarge, the SARI values dropped by 1 point on average compared to using the S4S training set alone. Besides, when we only used the WikiLarge training set, the SARI values dropped by an average of more than 10 points. We also gave example outputs in Table \ref{tab:Example2}. Above all, we believe the quality of the S4S dataset is higher than that of the Wikipedia-based datasets. The S4S dataset was 
given in the supplementary materials.

\subsection{Results on Extremely Low-resource Scenarios}

\begin{figure}[h]
\centering
\includegraphics[width=7.5cm]{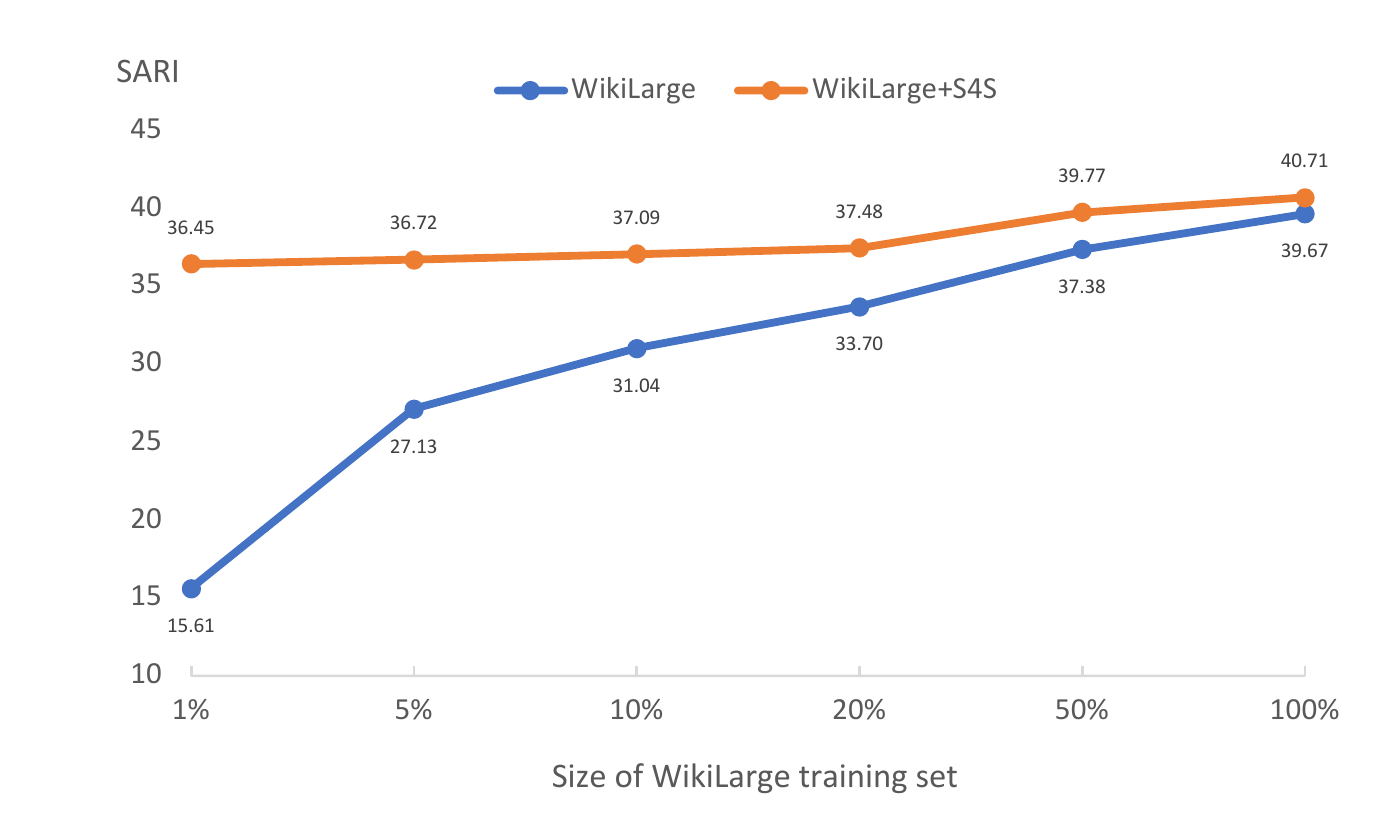}
\caption{Experimental results of extremely low-resource experiments on Turkcorpus test set.}
\label{pic:low_resource}
\end{figure}

In many cases simplification data is hard to obtain \cite{aprosio2019neural,maruyama2019extremely}, and we took a small amount of sentence pairs from the training set of WikiLarge to simulate an extremely low-resource situation. We reduced the size of the WikiLarge training set to 50\%, 20\%, 10\%, 5\%, and 1\%, respectively. We then conducted experiments using the ACCESS model trained on the size-reduced WikiLarge data and the mixture of size-reduced WikiLarge and S4S. The results are shown in Figure \ref{pic:low_resource}.

When the size of the training set is relatively small (less than 20\%, about 60,000 sentence pairs), S4S can improve the results significantly. The results prove that the S4S is effective in helping text simplification when data is difficult to obtain.

\subsection{Ablation Study}

In our proposed sentence filtering method, we used four attributes to control the simplicity of the sentence pairs extracted from summarization datasets. We removed the attributes one by one and then used the remaining three attributes as new rules to filter simple sentence pairs. We set ${\rm T_{s}}$ to 2.75 in the experiment. The filtered sentence pairs are mixed with the WikiLarge training set and then used to train the ACCESS model.

\begin{table}[h!]
\small
\centering
\renewcommand\arraystretch{1.12}{
\begin{tabular}{l|c}
\hline
Experiment                  & SARI$\uparrow$    \\ \hline
WikiLarge+S4S & 40.71        \\
WikiLarge            & 39.67        \\ \hline
Without word complexity             & 39.32(-1.39) \\
Without sentence length              & 39.63(-1.08) \\
Without word frequency                & 37.70(\textbf{-3.01}) \\
Without SARI value                & 38.78(-1.93) \\ \hline
\end{tabular}}
\caption{Ablation study on Turkcorpus test set.}
\label{tab:ablation}
\end{table}

The results are illustrated in Table \ref{tab:ablation}. In this experiment, the odds ratio attribute has the greatest effect on the results. When this attribute is missing, the SARI value decreases by 3.01 points. The sentence length attribute has the least effect on the results. When this attribute is missing, the SARI value drops by 1.08 points. The results also show that the four attributes of our design are meaningful. They all play a significant role in filtering the simplified sentence pairs.

\section{Conclusion}

In this paper, we are committed to mining data from text summarization datasets to help text simplification. We proposed an alignment algorithm and a new method to filter suitable sentence pairs. We named these pairs Sum4Simp (S4S). We conducted human evaluations on S4S and performed experiments on mainstream simplification models to illustrate that the S4S is high-quality and can help text simplification. In future work, we will apply our method to mine more simplification data from other summarization datasets.

\section*{Acknowledgements}
This work was supported by National Key R\&D Program of China (2021YFF0901502), National Science Foundation of China (No. 62161160339), State Key Laboratory of Media Convergence Production Technology and Systems and Key Laboratory of Science, Technology and Standard in Press Industry (Key Laboratory of Intelligent Press Media Technology). We appreciate the anonymous reviewers for their helpful comments. Xiaojun Wan is the corresponding author.




\section*{Limitations}
We considered the consumption of computational resources as the major limitation of our method. To extract aligned sentence pairs from summarization datasets, we need to calculate the similarity between each sentence in the summary and each sentence in the document, which makes the time complexity of the alignment algorithm be $O(n^2)$. We ran the alignment algorithm with an Intel Xeon processor. On average, there are 40 sentences in a document and 4 sentences in a summary. There are 312K documents in total with corresponding summaries. The total running time is 42,153s. We have released the aligned sentence pairs to help future research.

Second, to calculate the SARI values in Section \ref{sec:four_attributes}, we need to train a simplification model in advance, which can consume GPU resources. For example, if we train a BART model on the WikiLarge dataset and set the max epochs to 10, the training time spent on an Nvidia A40 is about 3 hours.

\bibliography{custom}
\bibliographystyle{acl_natbib}

\appendix

\section{More Details}
\label{sec:appendix}

In Algorithm \ref{alg:alignment}, for $S_{max}$, $S_{add}$, and $S_{min}$, we first observed the alignment results to obtain a rough range [0.5,0.8]. In this range, we set the step size to 0.1 and then chose four combinations of parameters: (0.8, 0.7, 0.6), (0.8, 0.7, 0.5), (0.8, 0.6, 0.5), and (0.7, 0.6, 0.5). We used human evaluation on 50 sentence pairs for each combination to determine which combination is the best. Finally, we set $S_{max}$ to 0.8, $S_{min}$ to 0.6, and $S_{add}$ to 0.7. Lmax is set to 3 as an empirical value. If it is too large, the model will be more concerned with deletion than simplification; if it is too small, the information in the original sentences will lose.

For the method of filtering suitable sentence pairs in Section \ref{sec:filtering_method}, we set $\alpha_{i}$ to 0.25 because it is difficult to prove that one of the four attributes is more important than the other. We performed a parameter research for ${\rm T_{s}}$ from 3.5 to 3.8 with a step size of 0.05. 

We have released the aligned sentence pairs obtained in Section \ref{sec:alignment_algorithm} for future research. So future researchers only need to set ${\rm T_{s}}$ when conducting experiments.

To obtain Table \ref{tab:expand_datasets}, we first trained models with existing simplification datasets (e.g., train ACCESS with WikiLarge). Then, we selected the model that performed best on the validation set to calculate the score $t$ for the SARI value attribute mentioned in Section \ref{sec:four_attributes}. In this way, we got S4S. We then trained models with WikiLarge+S4S to obtain the results in the fourth column of the Table \ref{tab:expand_datasets}. The S4S dataset in Section \ref{sec:real_simp_dataset} is obtained after we first trained ACCESS with WikiLarge. We will also release this version of S4S as a standard simplification dataset.

\begin{table}[h!]
\centering
\resizebox{0.48\textwidth}{!}{
\begin{tabular}{rl|rl}
\hline
\textbf{Parameter} & \textbf{Value} & \textbf{Parameter} & \textbf{Value} \\ \hline
epochs             & 30             & max source length  & 256            \\
batchsize          & 64             & max target length  & 256            \\
optimizer          & Adam           & dropout            & 0.1            \\
learning rate      & 5e-5           & $d_{model}$        & 768            \\
warm up steps      & 2000           & attention heads    & 12             \\ \hline
\end{tabular}}
\caption{Parameters of the Transformer model.}
\label{tab:parameter_transformer}
\end{table}

\begin{table}[h!]
\centering
\resizebox{0.48\textwidth}{!}{
\begin{tabular}{rl|rl}
\hline
\textbf{Parameter} & \textbf{Value} & \textbf{Parameter} & \textbf{Value} \\ \hline
epochs             & 10             & max source length  & 256            \\
batchsize          & 64             & max target length  & 256            \\
optimizer          & Adam           & dropout            & 0.1            \\
learning rate      & 5e-5           & $d_{model}$        & 768            \\
warm up steps      & 2000           & attention heads    & 12             \\ \hline
\end{tabular}}
\caption{Parameters of the BART model.}
\label{tab:parameter_bart}
\end{table}

\begin{table}[h!]
\centering
\resizebox{0.48\textwidth}{!}{
\begin{tabular}{rl|rl}
\hline
\textbf{Parameter} & \textbf{Value} & \textbf{Parameter} & \textbf{Value} \\ \hline
max epochs         & 100            & label smoothing    & 0.54           \\
max tokens         & 5000           & clip norm          & 0.1            \\
optimizer          & Adam           & dropout            & 0.2            \\
learning rate      & 1.1e-4         & weight decay       & 1e-4           \\
warm up steps      & 1000           & attention heads    & 8              \\ \hline
\end{tabular}}
\caption{Parameters of the ACCESS model.}
\label{tab:parameter_access}
\end{table}





\section{Definition of Alignment Quality}

\begin{table*}[h]
\centering
\small
\resizebox{15cm}{!}{
\renewcommand\arraystretch{1.12}{
\begin{tabular}{|l|l|}
\hline
Good &
  The semantics of the source sentence and the target sentence completely match, possibly with small omissions. \\ \hline
Source &
  Sets in children 's bedrooms or left on as background noise could be particularly damaging. \\
Target &
  Devices in bedrooms or left on as background noise is more damaging. \\ \hline
Good partial &
  \begin{tabular}[c]{@{}l@{}}Source and target sentence mean basically the same thing. However, source or target sentence may contain \\ additional information that is not contained in the other sentence.\end{tabular} \\ \hline
Source &
  \begin{tabular}[c]{@{}l@{}}The tape was played at a hearing Monday to determine whether or not the confession can be used as evidence\\ at Hernandez 's murder trial - not whether the statements are true. Judge Maxwell Wiley must decide whether\\ Hernandez was properly advised of his rights.\end{tabular} \\
Target &
  \begin{tabular}[c]{@{}l@{}}The judge must decide not whether the confession is true, but whether it can be permitted to be used as\\ evidence at Hernandez 's murder trial.\end{tabular} \\ \hline
Partial &
  \begin{tabular}[c]{@{}l@{}}Source and target sentence are discussing two unrelated concepts, but share short related phrases that do not\\ match considerably.\end{tabular} \\ \hline
Source &
  \begin{tabular}[c]{@{}l@{}}A non-profit group called Women On 20s, formed to convince President Barack Obama to put a woman’s \\ image on the \$20 note, already has done some polling.\end{tabular} \\
Target &
  There is a group called Women On 20s. \\ \hline
Bad &
  Source and target sentence are discussing two unrelated concepts. \\ \hline
Source &
  Leicester City have lost just one of their last seven league meetings with Hull City. \\
Target &
  88 \% of British grandmothers consider themselves to be a Glam-Ma. \\ \hline
\end{tabular}}}
\caption{Definition of the alignment quality. Example of each level of quality is also given.}
\label{tab:defination_of_alignment_quality}
\end{table*}

\label{sec:alignment_defination_and_examples}

\section{Example Outputs}

\begin{table*}[h]
\centering
\resizebox{15cm}{!}{
\renewcommand\arraystretch{1.12}{
\begin{tabular}{l|l}
\hline
Complex(input)    & in computing , a protocol is a set of rules which is used by computers to communicate with each other across a network . \\
Simple(reference) & in computing , a protocol is the language used by computers while talking with each other .                              \\ \hline
WikiSmall         & in computing , a protocol is a set of rules which is used by computers to communicate with each other across a network . \\
S4S               & the process is a set of rules which is used by computers to communicate with each other across a network                 \\
WikiSmall+OA      & in computing , a protocol is a set of rules which is used by computers to provide with each other across a network .     \\
WikiSmall+S4S     & in computing , a protocol is used by computers to communicate with each other across a network .  \\
\hline
\end{tabular}}}
\caption{An example of sentences generated by ACCESS. When the training set is WikiSmall, the complex sentence is not simplified. When the training set is S4S or WikiSmall+OA, the generated sentences contain grammatical errors and change the meaning of the complex sentence. The sentence generated by ACCESS trained on WikiSmall+S4S can be regarded as a simplified sentence.}
\label{tab:Example1}
\end{table*}

\begin{table*}[h]
\centering
\resizebox{15cm}{!}{
\renewcommand\arraystretch{1.12}{
\begin{tabular}{l|l}
\hline
Complex(input)    & barcelona manager luis enrique -lrb- pictured -rrb- insisted afterwards he was right to start uruguay striker suarez \\
Simple(reference) & barcelona boss luis enrique says he was right to start the uruguay player                                            \\ \hline
S4S               & barcelona boss luis enrique says he was right to start uruguay striker                                               \\
WikiLarge         & barcelona manager luis enrique -lrb- pictured - pictured he wanted to start uruguay striker suarez .                 \\
S4S+WikiLarge     & barcelona manager luis enrique said he was right to start right to start uruguay suarez suarez                       \\ \hline
\end{tabular}}}
\caption{An example of sentences generated by ACCESS when S4S is regarded as a standard simplification dataset. When the training set is WikiLarge, the generated sentence contains grammatical errors and changes the meaning of the complex sentence. When the training set is S4S+WikiLarge, the generated sentence also contains grammatical errors and is less simple than the generated sentence when the training set is S4S only. This example illustrates that the quality of S4S is higher than that of WikiLarge.}
\label{tab:Example2}
\end{table*}

\end{document}